# Modeling Teacher-Student Techniques in Deep Neural Networks for Knowledge Distillation

Sajjad Abbasi, Mohsen Hajabdollahi, Nader Karimi, Shadrokh Samavi
*Department of Electrical and Computer Engineering*
*Isfahan University of Technology Isfahan 84156-83111, Iran*

*Abstract*— Knowledge distillation (KD) is a new method for transferring knowledge of a structure under training to another one. The typical application of KD is in the form of learning a small model (named as a student) by soft labels produced by a complex model (named as a teacher). Due to the novel idea introduced in KD, recently, its notion is used in different methods such as compression and processes that are going to enhance the model accuracy. Although different techniques are proposed in the area of KD, there is a lack of a model to generalize KD techniques. In this paper, various studies in the scope of KD are investigated and analyzed to build a general model for KD. All the methods and techniques in KD can be summarized through the proposed model. By utilizing the proposed model, different methods in KD are better investigated and explored. The advantages and disadvantages of different approaches in KD can be better understood and develop a new strategy for KD can be possible. Using the proposed model, different KD methods are represented in an abstract view.

## I. Introduction

Knowledge Distillation (KD) is a particular type of knowledge transfer that has been developed in recent years. The main idea of KD is including two network structures, which are named teacher and student. The teacher is a model with a strong capability, while the student can be a simple model. The teacher model is used to teach the student model by transferring significant knowledge to the student [1].

The teacher-student (TS) model based on KD has been employed in several types of applications. Complexity and model of the student and teacher varied for different applications. Generally, there are three principal situations which TS models can be applied as follow:
- Design a model with a constraint in computational and power resources.
- Enhance network accuracy without imposing more complexity on its model.
- Train a model with constrained or limited training samples.

The most contributions conducted in the concept of TS model, are aiming to design a relatively small model that could be implemented in conditions with limited computational resources like mobile and portable devices. By employing these contributions, it is possible to substitute a cumbersome teacher with a small student model with acceptable accuracy [1]–[4].

Another contribution in the area of KD is to train a model based on a single teacher model with accuracies higher than the teacher, by distilling knowledge form an ensemble of teachers [1], [3], [5], [6]. Also, KD is suitable in situations that a limited number of training samples are available. In this situation, a trained model can be used to train another model with a small amount of data [1],[7].

Although the TS model has several useful applications, sometimes the employment of this model becomes complicated. By observing different studies, it is not possible to imagine a unique solution for solving related problems. In other words, there is not a universal and comprehensive routine to utilize different techniques and tricks available in KD.

In this study, we propose a comprehensive model containing different stages of utilizing TS based on KD. This model can be useful for designing a method based on TS from scratch. To the best of our knowledge, the current study is the first one attempting to consider various techniques conducted in the TS area in a single model. This model can be a guideline for the designer to identify the advantages and disadvantages of the different methods in the field of knowledge distillation.

The general model for the studies conducted based on the TS model is illustrated in Fig. 1. There are four main steps in the proposed model, as shown in Fig. 1. As illustrated in Fig. 1, any method utilizing the TS model has two principal inputs, including objective and data. Using the TS model, different purposes, such as simplicity and accuracy, can be achieved. Here TS models in the area of neural networks are investigated. Hence, the evaluation of a network can be possible by training and testing on the appropriate data.

All the transactions on the input data are performed in the first stage of the proposed model named data augmentation, as illustrated in Fig. 1. In some applications, networks of TS model, including teacher(s) or student(s), cannot be trained directly on the raw input data. Hence some transformation is required. This transformation can be related to the objective of the model, as well as the number of teachers that are used.

In the second stage named teacher modeling, the general structure of a teacher is determined based on the model objectives. Teachers can be in the form of an ensemble of the networks working on different inputs. This stage is considered to determine the number of teachers, their architectures as well as the ways they communicate with each other.

Knowledge distillation is the most crucial stage in the proposed model, which specifies how the distillation is conducted. In the student modeling stage, the architecture of student(s), and the method of getting knowledge from the teacher(s) are considered. As it is shown in the Fig.1, this stage is directly connected to data & objective stage, because it is designed according to the type of problem and data.

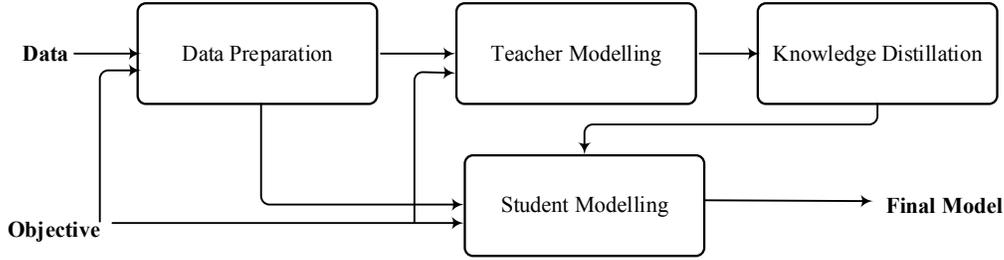

Fig.1. The proposed model for the studies conducted in the area of TS model by KD

In the following sections, the details of stages illustrated in Fig. 1 are presented. The remaining part of this study is organized as follows. In Section II, we investigate different approaches based on their objectives and data types. In Section III and IV, different methods for modeling of teachers and students are presented, respectively. In Section V, knowledge distillation methods are modeled and summarized. In Section VI, the results of the summarization of different methods under the proposed model are presented. Finally, in Section VII, concluding remarks are presented.

## II. OBJECTIVES & DATA

Several objectives can be expected from a problem in the area of the TS model using KD. Objectives are fundamental to form an appropriate solution for the problem. The most common objectives that are considered in the TS model are including design a simplified model, design a model in situations with limited data access, short-time learning requirement, and design a model with better accuracy.

In Fig. 2, objectives expected from employing the TS model are illustrated. Four goals are considered in previous studies. Smartphones are one of the most handheld devices that have a constraint in their resources. Many studies in the TS model have developed a model that works on smartphones. In [5], [2], [4], and [8] the problem of design a simple model with a low memory requirement is addressed. Sometimes in medical image processing, there is a problem of data access limitation and employing the proposed TS model can be very beneficial. In [1], [7] and [9] small parts of input data are used to train a student model. Moreover, utilizing a small amount of data for training reduces the learning time. Also, training a model with a large number of input data like JFT can be a very time-consuming process. In this regard, in [1], a method is developed for parallel training the student model to reduce to total training time. Improving model accuracy can be regarded as another objective in the TS model area. In [1], [10], [11], [9] and [6] the principal objectives are improving the accuracy of the teacher(s) by training specialized students.

In addition to objectives, data is another input of the proposed TS model. Data illustrated in Fig.1 means different types of inputs which are suitable for training teacher and student models. Providing appropriate data for the model has played an essential role in having an efficient TS model. In some conditions, different types of data are fed to the teacher and student. An example of the importance of data providing is presented by Hinton et al. in [1], in which a fast model is designed. In their work to create a fast model, the complete dataset is used for training teachers, but just 3% of the dataset is considered for the training of the student model.

## III. DATA PREPARATION

Input data often cannot be added directly to the TS model. So a data preparation including data transform and model extraction as well as data packaging should be performed. Data preparation steps including data modification and data partitioning, are illustrated in Fig. 3. In data modification, according to the objective of the model, input data are transformed and in some cases, different types of input data are employed. Various types of data can be provided by different transformation and acquisition. Mahbod et al.[12] proposed a method in which some fundamental augmentations like rotation and horizontal flipping are applied on the training data. Wu et al.[5] suggested a method in which different views of data are employed. In the first experiment of this work, I-frames, motion vectors and residuals are considered as inputs of the teacher while just I-frame is considered for student model [5]. Zhu et al.[6] proposed a new manner in which the input data, CIFAR-10 images, are down-sampled to 8×8 or 16×16 and then up-sampled to the original size of 32×32. The generated data are used to train the same teacher and student model. In [1], input data is partitioned into different parts and used to train the teacher models. In [13], the input data is fed to the TS model in the form of four different types of MRI images including T1W, T2W, DWI, and DYN. Each type of image is used to train a separated teacher model. From Fig. 3, it can be observed that data modification could be in the form of different resolutions, different augmentations, different modalities, and different views. Also, partitioning based on the data transformation can be for teacher(s), student(s), and both of them [5],[6].

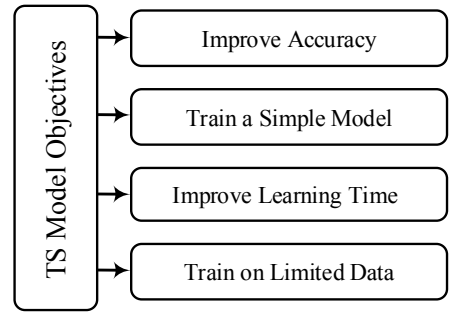

Fig. 2. Objectives in TS model

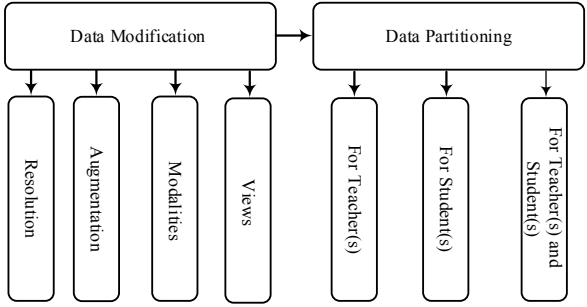

Fig .3. Data preparation.

## IV. TEACHER MODELING

Teacher modeling is the main stage of the proposed model in which teacher specifications are described. All the techniques about teacher designing are summarized in this stage. According to the previous studies, the teacher modeling stage includes three steps as illustrated in Fig. 4. The first one is the teacher model setup. The number of teachers working on different input types is determined in the teacher model setup. There are various methods for setup a teacher model. In [6], the teacher is a single baseline model that is trained on images of the CIFAR-10 dataset. The student is also a single model trained on a different resolution of input images. In [1], the teacher is an ensemble of 10 similar baseline networks and the student is a single model trained on the aggregation of their results. In [5], the teacher model is the ensemble of 3 different structures. One of the teacher structure is trained on I-frame data while and others are trained on residual and motion vectors. In [12], three teachers with different structures that are pre-trained on Image-Net are used in parallel to extract deep features of images.

The next step of the teacher modeling stage is the selection of architecture for the teacher. In the architecture selection step, the concentration is on selecting suitable architecture for the teacher(s). Structures with different complexities can be used for teacher model. In this step, the type of input data and the quantity and complexity of data partitions should be considered. In [8], the TS model is designed as a neural machine translation (NMT). The input data of the NMTs are in an extensive sequence, so an LSTM is designed for teacher structure. In [5] for the I-frame data partitions, a ResNet-152 is dedicated, which is a complex structure proportional to the complexity of analysis of the I-frames. On the other hand, for analysis of residual and motion vector partitions, which is more straightforward than the analysis of I-frames, two ResNet-18 structures are considered separately.

In Teacher Construction & Train stage, the primary consideration is about how to train the teacher model and transfer knowledge, as illustrated in Fig. 4. When there are multi-teachers, teachers can be trained in different ways. Also, knowledge distillation can be performed in different ways, depending on the application. In some implementations, training the teacher model is not in a straightforward way. For example, in [4], the teacher is trained with low-precision features to have a student model with low-precision.

## V. KNOWLEDGE DISTILLATION

Assume training data tuples of inputs and labels $(x.y) \in D$ which D is a set of training data. Let T be Teacher network with parameters $\theta_T$ and S be Student network with parameters $\theta_S$. The distillation equation can be written as Eq. (1) to minimize

$$L = \sum_{(x.y)\in D} L_{KD}\big(S(x.\theta_S.\gamma).T(x.\theta_T.\gamma)\big) + \varphi L_{CE}(\hat{y}_S.y) \quad (1)$$

$L$.

In Eq. (1), $L_{CE}$ is cross-entropy which is computed on the labels $\hat{y}_S$ that are predicted by the student and ground truth labels y with temperature =1. $L_{KD}$ is distillation loss, which is cross-entropy computed on softmax output of teacher and student with temperature $\gamma$. Hence, $S(x.\theta_S.\gamma) \text{ and } T(x.\theta_T.\gamma)$ represent of softmax output of the student and the teacher, respectively. $\varphi$ is hyper-parameters to balance the influence of each loss [7][1].

How the knowledge is transferred between the teacher(s) and student(s) is specified in this stage. As illustrated in Fig. 5, there are three steps in knowledge distillation playing an essential role in having an efficient TS model. These steps are including the determination of knowledge types, location of distillation, and methods of knowledge transfer.

In the base TS models, soft-labels (also known as logits) are considered as distilled knowledge. However, the knowledge can be distilled from each location of the teacher model, including the end of the model and between layers. In [14], knowledge is transferred between blocks of the teacher to student. Distillation-loss is realized through a cross-entropy function that is applied to the output of the student and soft-labels of the teacher. Different knowledge types are considered in previous studies, such as soft labels, hard labels, etc. In [9], knowledge type in the form of the mutual information between intermediate layers is maximized and several functions are used to minimize the loss of intermediate layers. In [7], knowledge

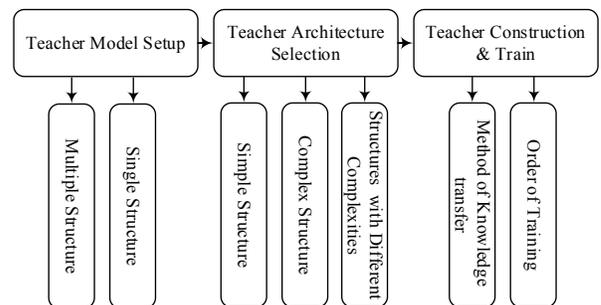

Fig. 4. Teacher modeling

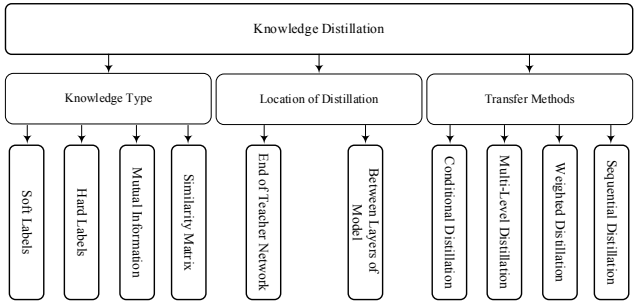

Fig. 5. Knowledge distillation techniques

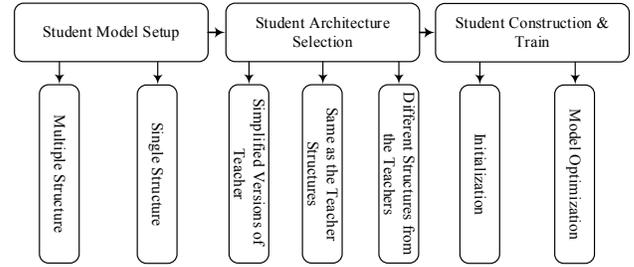

Fig.6. Student modeling

types are in the form of a similarity matrix between teacher and student models.

Transfer Methods is another step in the knowledge distillation stage. In [15] a multilevel approach for knowledge transfer is presented in which an information mask is provided by the teacher, and the student is trained with both of the information masks and teacher's ground truth. Also, in [16] and [17], there is more than one level of distillation. In [16], student is trained in three levels of distilled information including segmenting, classification, and semantic segmentation. Also, in [17], there are two levels of knowledge distillation considering the information between teachers as well as information between teachers and students. In [10], a conditional method is employed in which a controller is applied such that if teacher prediction is false, the student trains with the original labels. In [11], a weighted approach is utilized. Teachers assign input data a weight corresponding to confidence about their prediction. Also, in [18], a method is introduced emphasizing the training samples, which have more contribution to the teacher training.

## VI. STUDENT MODELLING

The final stage of TS modeling is the modeling of the student. All the specifications of a student(s) should be specified in this stage. The output of this stage is regarded as the final result of a TS model. As illustrated in Fig. 6, this stage has three steps including Student Model Setup, Student Architecture Selection, and Student Construction & Train. The main contents of this stage are similar to the teacher model. However, among the teacher and student, only student model works during inference time and should be corresponded to the final objectives.

As depicted in Fig. 6, in student model setup, the number of students, to have the ability of appropriate knowledge transfer, is considered. In [8], there are several students that each one is like the corresponded teachers. In [5], three students are trained on three different data partitions. Generally, the student model is a simpler model than the teacher, but there are models that student is more complex [2].

In the architecture selection, the appropriate structure is determined for students. The architecture of student and teacher can be the same as stated in [1], [6] and [10]. In some applications, the student has a small structure, which is a simplified version of the teacher, as stated in [5], [4], [2], [7] and [9]. In some experiments, the architecture of the student and teacher is different. In [9] in one of the conducted experiments, a ResNet-34 is considered as a teacher, and a VGG-9 is considered as a student. Summarization of the Student Architecture Selection techniques is illustrated in Fig. 6. It can be observed that a student model can be in the form of the teacher model, a simplified version of the teacher, and different from the teacher.

In Student Construction and Train step, training a student model with its objectives is considered. Different model optimizations can be applied in this step. In paper [2], a quantized student is trained on distilled data of the teacher. This quantization is caused by significant compression and satisfies the objective of the model. In [4], both teacher and student are defined with low-precision parameters. Also, in [8], knowledge of teachers is used to train a pruned student. Although the student can be trained from scratch, by initializing the student with teacher parameters, better training is possible [1].

## VII. RESULTS OF SUMMARIZING DIFFERENT STUDIES UNDER THE PROPOSED MODEL

By viewing the proposed general structure, the summarization of different methods in the TS model becomes possible. In Table I, and Table II, the results of summarizing different methods under the proposed model are illustrated. Based on the results of Table I and Table II, three main objectives are widely addressed in the previous studies.

### A. Accuracy Enhancement

It can be observed from Table I and Table II, that the accuracy can be improved by employing the following techniques.

*1) Multi-Structure:* Using multiple specialists, which are trained on separated parts of the dataset, led to a 4.4% relative improvement in test accuracy [1]. Equipping these specialists with KD increased the accuracy of 1.9% in the case of a student. In [11] employing multiple teachers leads to better accuracy. Also, using a series of students, which sequentially transform knowledge, decreased the test error on CIFAR-100 about 5% by using Res-Net architecture as baseline [11].

*2) Distillation Techniques:* During distillation, in samples that teacher prediction is not correct, using the one-hot labels leads to better accuracy of 1.78% than the conventional distillation[10]. Also, weighted knowledge causes better results in comparison with conventional distillation [11].

## B. Model Size Reduction

A reduction in model size can be realized by different techniques as follows.

*1) Multi-Teacher structure:* In video action recognition, by using three teachers with I-frame, motion vector and residual as their input data, a better compression rate of 2.4 with only 1.79% drop of accuracy is obtained [5]**.**

*2) Student Optimization:* By using a TS model such that the student and teacher are designed with *low-precision* parameters, it is possible to obtain significant computational reduction. With ResNet-101 architecture as teacher and ResNet-18 as a student, computations are reduced significantly [4]. Also, applying pruning on the student makes it significantly compressed. In [8], a student with 80% parameter pruning and a slight drop in accuracy results.

*3) Distillation method:* In [2], by initializing the student with the quantized weights of the teacher model, the student can be compressed significantly. Also, considering 32 bits full precision CNN as a teacher and an 8 bits precision CNN as a student, it is possible to even have 0.02% better accuracy than the full precision TS model in CIFAR-10 dataset [2].

## C. Learning with Limited data

Distillation methods can deal with problems in which data is limited, or data access is restricted. In [9], by distilling the knowledge between intermediate layers, the accuracy of the TS model is not decreased with fewer input data. In another experiment with the same structure, using just 2% of input data for training would result in a 10.26% accuracy drop [9].

As stated before, in some conditions using the original data for training the student is not possible. So using the impression of data can be very useful [7]. Data Impression is the modified similarity matrix of the weights in the last layer of the teacher model. Generally, it can be observed from Table I and Table II that modern techniques are concentrating on using the multiple and different structures each one working on different types of input data. With multiple structures, the ways of knowledge distillation between different structures play an essential role to have an efficient TS model. Also, types of knowledge under distillation are very important because appropriate data should be provided for the corresponding task.

## VIII. CONCLUDING REMARKS

A general structure to model the studies conducted in the teacher-student paradigm through knowledge distillation was proposed. Four main stages were considered for the TS modeling, and all the studies were summarized under that. Better insight and understanding of different methods can be possible using the proposed model. In designing a TS model, three main cases should be designed efficiently. First, the structures of the teacher and student. Second, the type of the transferred knowledge, and third, the method of transferring of the knowledge. Summarizing different methods using the proposed model, we can indicate that using multiple teachers and students as well as using an abundant knowledge yield the best model accuracy. Also, concentrating on the student model and a suitable initialization leads to a model with appropriate simplicity.

TABLE I. Summarization of different methods on TS based on the proposed general model.

| Work | Objective | Data Modification | Teacher Model | Knowledge Type | KD method | Student Model | Dataset |
|---|---|---|---|---|---|---|---|
| [6] | Accuracy | For student, Down-sampled images | Single, Simple and complex | Soft labels | Permuted Distillation | Single, Teacher-like | CIFAR10, SVHN |
| [2] | simple model | - | Single, Simple and complex | Soft labels | Permuted Distillation | Single, Quantized CNN | CIFAR10, WMT13, CIFAR100, OPENNMT |
| [11] | Accuracy | - | Single, Complex | Soft labels | Sequential, Permuted Distillation | Single, Teacher-like | CIFAR10, CIFAR100, PTB |
| [7] | limited data | Make Data Impression | Single, Simple | Soft labels | Permuted Distillation | Single, Teacher-like | MNIST, FMNIST, CIFAR10 |
| [1] (Ex.1) | Accuracy | - | Multi-teacher | Soft labels | Aggregated Distillation | ASR | JFT |
| [1] (Ex.2) | limited data | Decrease training data | Multi-teacher | Soft labels | Aggregated Distillation | Single, not special (CNN) | JFT |
| [4] | simple model | - | Single, Simple | Soft labels | Permuted Distillation | Single, Teacher-like | TINY IMAGENET |
| [5] | simple model | Extract motion vector and residual | Multi-teacher | Soft labels | Aggregated Distillation | Multi-student, Teacher-like | UDF101, HMDB51 |
| [9] | limited data | Decrease training data | Single, Complex | Mutual information | Intermediate layers, Permuted Distillation | Single, Simplified teacher | CIFAR10, CIFAR100, MIT, CUB200 |

TABLE II. Summarization of different methods on TS based on the proposed general model.

| Work | Objective | Data Preparation | Teacher Model | Knowledge Type | KD method | Student Model | Dataset |
|---|---|---|---|---|---|---|---|
| [14] | Accuracy | Eliminate some parts of data for student | Single, Complex | Mutual information | Intermediate layers, Permuted Distillation | Single, Simplified teacher | SYSU 3D HOI, NTU RGB-D Action, UCF101 |
| [15] | simple model | - | Single, Complex | Mutual information | Permuted Distillation Multi-level | Single, Simplified teacher | COCO, PASCAL, KITTI |
| [16] | simple model | - | Single, Complex | Mutual information | Intermediate layers, Multi-level Distillation | Single, Simple | Cityscapes, CAMVID, ADE20K |
| [17] | simple model | Use mutual relation of data | Single, Complex | Soft labels | Permuted Distillation, Multi-level Distillation | Single, Simplified teacher | CUB200, Cars196, Stanford Online Product |
| [8] | learning time | - | Single, Complex | Mutual information | Intermediate layers, Permuted Distillation | Single, Simplified teacher | WMT 2014, IWSLT 2015 |
| [10] | Accuracy | - | Single, Complex | Soft labels | Conditional Distillation | Single, Teacher-like | CHiME-3, Microsoft Short Message Dictation |
| [13] | Accuracy | Extract 4 diff. imaging mode | Multi-teacher | Hard labels | Aggregated Distillation | Implicit student | MRI dataset |
| [12] | Accuracy | Fundamental Augmentations | Multi-teacher | Deep features | Permuted Distillation | Single, Complex(SVM) | ISIC 2017 |
| [18] | learning time and accuracy | - | Single, Complex | Soft labels | Weighted Distillation | Single, Simplified teacher | LIBRISPEECH |

REFERENCES

[1] G. Hinton, O. Vinyals, and J. Dean, "Distilling the knowledge in a neural network," *arXiv Prepr. arXiv1503.02531*, 2015.
[2] A. Polino, R. Pascanu, and D. Alistarh, "Model compression via distillation and quantization," *arXiv Prepr. arXiv1802.05668*, 2018.
[3] R. Anil, G. Pereyra, A. Passos, R. Ormandi, G. E. Dahl, and G. E. Hinton, "Large scale distributed neural network training through online distillation," *arXiv Prepr. arXiv1804.03235*, 2018.
[4] A. Mishra and D. Marr, "Apprentice: Using knowledge distillation techniques to improve low-precision network accuracy," *arXiv Prepr. arXiv1711.05852*, 2017.
[5] M.-C. Wu, C.-T. Chiu, and K.-H. Wu, "Multi-teacher Knowledge Distillation for Compressed Video Action Recognition on Deep Neural Networks," *IEEE International Conference on Acoustics, Speech and Signal Processing (ICASSP)*, pp. 2202–2206, 2019.
[6] M. Zhu, K. Han, C. Zhang, J. Lin, and Y. Wang, "Low-resolution Visual Recognition via Deep Feature Distillation," *IEEE International Conference on Acoustics, Speech and Signal Processing (ICASSP)*, pp. 3762–3766, 2019
[7] G. K. Nayak, K. R. Mopuri, V. Shaj, R. V. Babu, and A. Chakraborty, "Zero-Shot Knowledge Distillation in Deep Networks," *arXiv Prepr. arXiv1905.08114*, 2019.
[8] Y. Kim and A. M. Rush, "Sequence-level knowledge distillation," *arXiv Prepr. arXiv1606.07947*, 2016.
[9] S. Ahn, S. X. Hu, A. Damianou, N. D. Lawrence, and Z. Dai, "Variational information distillation for knowledge transfer," *IEEE Conference on Computer Vision and Pattern Recognition*, pp. 9163–9171, 2019
[10] Z. Meng, J. Li, Y. Zhao, and Y. Gong, "Conditional teacher-student learning," *IEEE International Conference on Acoustics, Speech and Signal Processing (ICASSP)*, pp. 6445–6449, 2019.
[11] T. Furlanello, Z. C. Lipton, M. Tschannen, L. Itti, and A. Anandkumar, "Born again neural networks," *arXiv Prepr. arXiv1805.04770*, 2018.
[12] A. Mahbod, G. Schaefer, C. Wang, R. Ecker, and I. Ellinge, "Skin lesion classification using hybrid deep neural networks," *IEEE International Conference on Acoustics, Speech and Signal Processing (ICASSP)*, pp. 1229–1233, 2019
[13] W. Lu, Z. Wang, Y. He, H. Yu, N. Xiong, and J. Wei, "Breast Cancer Detection Based on Merging Four Modes Mri Using Convolutional Neural Networks," *IEEE International Conference on Acoustics, Speech and Signal Processing (ICASSP)*, pp. 1035–1039, 2019.
[14] X. Wang, J.-F. Hu, J.-H. Lai, J. Zhang, and W.-S. Zheng, "Progressive Teacher-student Learning for Early Action Prediction," *IEEE Conference on Computer Vision and Pattern Recognition*, pp. 3556–3565, 2019.
[15] T. Wang, L. Yuan, X. Zhang, and J. Feng, "Distilling Object Detectors with Fine-grained Feature Imitation," in *Proceedings of the IEEE Conference on Computer Vision and Pattern Recognition*, pp. 4933–4942, 2019.
[16] Y. Liu, K. Chen, C. Liu, Z. Qin, Z. Luo, and J. Wang, "Structured Knowledge Distillation for Semantic Segmentation," in *Proceedings of the IEEE Conference on Computer Vision and Pattern Recognition*, pp. 2604–2613, 2019.
[17] W. Park, D. Kim, Y. Lu, and M. Cho, "Relational Knowledge Distillation," in *Proceedings of the IEEE Conference on Computer Vision and Pattern Recognition*, pp. 3967–3976, 2019.
[18] H.-G. Kim *et al.*, "Knowledge Distillation Using Output Errors for Self-attention End-to-end Models," *IEEE International Conference on Acoustics, Speech and Signal Processing (ICASSP)*, pp. 6181–6185, 2019.